\documentclass[a4paper,11pt]{article}

\usepackage[english]{babel}
\usepackage[utf8x]{inputenc}
\usepackage[T1]{fontenc}

\usepackage[a4paper,top=2.5cm,bottom=2.5cm,left=2.5cm,right=2.5cm,marginparwidth=1.75cm]{geometry}

\usepackage[colorlinks=true, allcolors=blue, backref=page]{hyperref}

\usepackage{amsmath}
\usepackage{mathtools}
\usepackage{graphicx}
\usepackage[colorinlistoftodos]{todonotes}
\usepackage{float}
\usepackage{url}
\usepackage{amsfonts}
\usepackage{caption}
\usepackage{hyperref}
\usepackage{nicefrac}
\usepackage{datetime}

\DeclarePairedDelimiter\abs{\lvert}{\rvert}%
\DeclarePairedDelimiter\norm{\lVert}{\rVert}%

\newcommand{\cev}[1]{\reflectbox{\ensuremath{\vec{\reflectbox{\ensuremath{#1}}}}}}

\makeatletter
\let\oldabs\abs
\def\abs{\@ifstar{\oldabs}{\oldabs*}}
\let\oldnorm\norm
\def\norm{\@ifstar{\oldnorm}{\oldnorm*}}
\makeatother

\newdateformat{monthyeardate}{%
  \monthname[\THEMONTH] \THEYEAR}

\usepackage{fancyhdr}
\hypersetup{pdfauthor={Antoine J.-P. Tixier},%
            pdftitle={Introduction to Deep Learning for NLP},%
            pdfsubject={},%
            pdfkeywords={Artificial Intelligence, Deep Learning, NLP, CNN, LSTM, Neural Network, Attention, seq2seq, encoder-decoder},%
            pdfproducer={LaTeX},%
            pdfcreator={pdfLaTeX}
}

\title{Notes on Deep Learning for NLP}
\author{Antoine J.-P. Tixier \\ 
Computer Science Department (DaSciM team)\\ 
\'Ecole Polytechnique, Palaiseau, France \\
\texttt{antoine.tixier-1@colorado.edu} }

\date{\small{Last updated \today~(first uploaded March 23, 2017)}}

\begin{document}
\maketitle

\tableofcontents

\section{Disclaimer}
Writing these notes is part of my learning process, so it is a work in progress. To write the current version of this document, I curated information mainly from the original 2D CNN paper \cite{lecun1998} and Stanford's \texttt{CS231n} CNN course notes\footnote{{\scriptsize\url{http://cs231n.github.io/convolutional-networks/}}}, Zhang and Wallace practitioners' guide to CNNs in NLP \cite{Zhang2015practicioner}, the seminal papers on CNN for text classification \cite{JohnsonZhang,kim2014}, Denny Britz' tutorial\footnote{{\scriptsize\url{http://www.wildml.com/2015/09/recurrent-neural-networks-tutorial-part-1-introduction-to-rnns/}}} on RNNs, Chris Colah's post\footnote{{\scriptsize\url{http://colah.github.io/posts/2015-08-Understanding-LSTMs/}}} on understanding the LSTM unit, and the seminal papers on the GRU unit \cite{gru,gru_emp}, encoder-decoder architectures \cite{gru,seq-to-seq} and attention \cite{att_luong, att_bahdanau}. Last but not least, Yoav Golderg's primer on neural networks for NLP \cite{goldberg2016primer} and Luong, Cho and Manning tutorial on neural machine translation\footnote{{\scriptsize\url{https://sites.google.com/site/acl16nmt/home}}} proved very useful.

\section{Code}
\noindent I implemented some of the models described in this document in Keras and tested them on the IMDB movie review dataset. The code can be found on my GitHub: \url{https://github.com/Tixierae/deep_learning_NLP}. Again, this is a work in progress.

\section{IMDB Movie review dataset}\label{sec:imdb}
\subsection{Overview}
The task is to perform binary classification (positive/negative) on reviews from the Internet Movie Database (IMDB) dataset\footnote{{\scriptsize\url{http://ai.stanford.edu/~amaas/data/sentiment/}}}, which is known as \textit{sentiment analysis} or \textit{opinion mining}. The dataset contains 50K movie reviews, labeled by polarity. The data are partitioned into 50 \% for training and 50\% for testing. The \texttt{imdb\_preprocess.py} script on my GitHub cleans the reviews and put them in a format suitable to be passed to neural networks: each review is a list of word indexes (integers) from a dictionary of size $V$ where the most frequent word has index 1.

\subsection{Binary classification objective function}
The objective function that our models will learn to \textit{minimize} is the \textit{log loss}, also known as the \textit{cross entropy}. More precisely, in a binary classification setting with 2 classes (say 0 and 1) the log loss is defined as:
\begin{equation}
\mathrm{logloss} = -\frac{1}{N}\sum_{i=1}^{N}\big(y_{i}\mathrm{log}p_{i} + \big(1- y_{i}\big)\mathrm{log}\big(1-p_{i}\big)\big)
\end{equation}

\noindent Where $N$ is the number of observations, $p_{i}$ is the probability assigned to class 1, $\big(1-p_{i}\big)$ is the probability assigned to class 0, and $y_{i}$ is the true label of the $i^{th}$ observation (0 or 1). You can see that only the term associated with the true label of each observation contributes to the overall score. For a given observation, assuming that the true label is 1, and the probability assigned by the model to that class is 0.8 (quite good prediction), the log loss will be equal to $-\mathrm{log}(0.8)=0.22$. If the prediction is slightly worse, but not completely off, say with $p_{i} = 0.6$, the log loss will be equal to $0.51$, and for $0.1$, the log loss will reach $2.3$. Thus, the further away the model gets from the truth, the greater it gets penalized. Obviously, a perfect prediction (probability of 1 for the right class) gets a null score.

\section{Paradigm switch}

\subsection{Feature embeddings}
Compared to traditional machine learning models that consider core features and combinations of them as unique dimensions of the feature space, deep learning models often \textit{embed} core features (and core features only) as vectors in a low-dimensional continuous space where dimensions represent shared latent concepts \cite{goldberg2016primer}. The embeddings are initialized randomly or obtained from pre-training\footnote{In NLP, pre-trained word vectors obtained with Word2vec or GloVe from very large corpora are often used. E.g., Google News \texttt{word2vec} vectors can be obtained from \url{https://code.google.com/archive/p/word2vec/}, under the section ``Pre-trained word and phrase vectors''}. They can then be updated during training just like other model parameters, or be kept static. 

\subsection{Benefits of feature embeddings}
The main advantage of mapping features to dense continuous vectors is the ability to capture similarity between features, and therefore to generalize. For instance, if the model has never seen the word ``Obama'' during training, but has encountered the  word ``president'', by knowing that the two words are related, it will be able to transfer what it has learned for ``president'' to cases where ``Obama'' is involved. With traditional one-hot vectors, those two features would be considered orthogonal and predictive power would not be able to be shared between them\footnote{Note that one-hot vectors can be passed as input to neural networks. But then, the network implicitly learns feature embeddings in its first layer}. Also, going from a huge sparse space to a dense and compact space reduces computational cost and the amount of data required to fit the model, since there are fewer parameters to learn.

\subsection{Combining core features}
Unlike what is done in traditional ML, combinations of core features are not encoded as new dimensions of the feature space, but as the \textit{sum}, \textit{average}, or \textit{concatenation} of the vectors of the core features that are to be combined. Summing or averaging is an easy way to always get a fixed-size input vector regardless of the size of the training example (e.g., number of words in the document). However, both of these approaches completely ignore the ordering of the features. For instance, under this setting, and using unigrams as features, the two sentences ``John is quicker than Mary'' and ``Mary is quicker than John'' have the exact same representation. On the other hand, using concatenation allows to keep track of ordering, but \textit{padding} and \textit{truncation}\footnote{\url{https://keras.io/preprocessing/sequence/}} need to be used so that the same number of vectors are concatenated for each training example. For instance, regardless of its size, every document in the collection can be transformed to have the same fixed length $s$: the longer documents are truncated to their first (or last, middle...) $s$ words, and the shorter documents are padded with a special zero vector to make up for the missing words \cite{Zhang2015practicioner,kim2014}.

\section{Convolutional Neural Networks (CNNs)}
\subsection{Local invariance and compositionality}
Initially inspired by studies of the cat's visual cortex \cite{hubel1962receptive}, CNNs were developed in computer vision to work on regular grids such as images \cite{lecun1998}. They are feedforward neural networks where each neuron in a layer receives input from a neighborhood of the neurons in the previous layer. Those neighborhoods, or \textit{local receptive fields}, allow CNNs to recognize more and more complex patterns in a hierarchical way, by combining lower-level, elementary features into higher-level features. This property is called \textit{compositionality}. For instance, edges can be inferred from raw pixels, edges can in turn be used to detect simple shapes, and finally shapes can be used to recognize objects. Furthermore, the absolute positions of the features in the image do not matter. Only capturing their respective positions is useful for composing higher-level patterns. So, the model should be able to detect a feature regardless of its position in the image. This property is called \textit{local invariance}. Compositionality and local invariance are the two key concepts of CNNs.

CNNs have reached very good performance in computer vision \cite{krizhevsky}, but it is not difficult to understand that thanks to compositionality and local invariance, they can also do very well in NLP. Indeed, in NLP, high-order features ($n$-grams) can be constructed from lower-order features just like in CV, and ordering is crucial locally (``not bad, quite good'', ``not good, quite bad'', ``do not recommend''), but not at the document level. Indeed, in trying to determine the polarity of a movie review, we don't really care whether ``not bad, quite good'' is found at the start or at the end of the document. We just need to capture the fact that ``not'' precedes ``bad'', and so forth. Note that CNNs are not able to encode long-range dependencies, and therefore, for some tasks like language modeling, where long-distance dependence matters, recurrent architectures such as LSTMs are preferred.

\subsection{Convolution and pooling}
Though recent work suggests that convolutional layers may directly be stacked on top of each other \cite{springenberg2014striving}, the elementary construct of the CNN is a \textit{convolution} layer followed by a \textit{pooling} layer. In what follows, we will detail how these two layers interplay, using as an example the NLP task of short document classification (see Fig. \ref{fig:cnn}).

\subsubsection{Input}
We can represent a document as a real matrix $A \in \mathbb{R}^{s \times d}$, where $s$ is the document length, and $d$ is the dimension of the word embedding vectors. Since $s$ must be fixed at the collection level but the documents are of different sizes, we truncate the longer documents to their first $s$ words, and pad the shorter documents with a special zero vector as many times as necessary. The word vectors may either be initialized randomly or be pre-trained. In the latter case, they can be updated during training or not (``non-static'' vs. ``static'' approach \cite{kim2014}). 

Thinking of $A$ as an image is misleading, because there is only one spatial dimension. The embedding vectors are not actually part of the input itself, they just represent the coordinates of the elements of the input in a shared latent space. In computer vision, the term \textit{channels} is often used to refer to this \textit{depth} dimension (not to be mistaken with the number of hidden layers in the network). If we were dealing with images, we would have two spatial dimensions, plus the depth. The input would be a tensor of dimensionality $(\mathrm{width} \times \mathrm{height} \times \mathrm{n\_channels})$, i.e., a 2D matrix where each entry would be associated with a vector of length 3 or 1, respectively in the case of color (RGB) and grey level images.

\subsubsection{Convolution layer}
The convolution layer is a linear operation followed by a nonlinear transformation. The linear operation consists in multiplying (elementwise) each instantiation of a 1D window applied over the input document by a \textit{filter}, represented as a matrix of parameters. The filter, just like the window, has only one spatial dimension, but it extends fully through the input depth (the $d$ dimensions of the word embedding space). If $h$ is the window size, the parameter matrix $W$ associated with the filter thus belongs to $\mathbb{R}^{h \times d}$. $W$ is initialized randomly and learned during training. 

The instantiations of the window over the input are called \textit{regions} or \textit{receptive fields}. There are $\nicefrac{(s−h)}{\mathrm{stride}}+1$ of them, where $\mathrm{stride}$ corresponds to the number of words by which we slide the window at each step. With a stride of 1, there are therefore $s-h+1$ receptive fields. The output of the convolution layer for a given filter is thus a vector $o \in \mathbb{R}^{s-h+1}$ whose elements are computed as:

\begin{equation}
o_{i} = W \cdot A[i:i+h-1,:]
\end{equation}
Where $A[i:i+h-1,:] \in \mathbb{R}^{h \times d}$ is the $i^{th}$ region matrix, , and $\cdot$ is an operator returning the sum of the row-wise dot product of two matrices. Note that for a given filter, the same $W$ is applied to all instantiations of the window regardless of their positions in the document. In other words, the parameters of the filter are shared across receptive fields. This is precisely what gives the spatial invariance property to the model, because the filter is trained to recognize a pattern wherever it is located. It also greatly reduces the total number of parameters of the model.


Then, a nonlinear activation function $f$, such as \texttt{ReLU}\footnote{compared to tanh, ReLu is affordable (sparsity induced by many zero values in the negative regime) and better combats the \textit{vanishing gradients} problem as in the positive regime, the gradient is constant, whereas with tanh it becomes increasingly small} ($\mathrm{max}(0,x)$) or \texttt{tanh} ($\frac{e^{2x}-1}{e^{2x}+1}$), is applied elementwise to $o$, returning what is known as the \textit{feature map}  $c \in \mathbb{R}^{s-h+1}$ associated with the filter:

\begin{equation}
c_{i} = f(o_{i}) + b
\end{equation}
Where $b \in \mathbb{R}$ is a trainable bias.\\

\begin{figure}[H]
\centering
\includegraphics[width=0.71\textwidth]{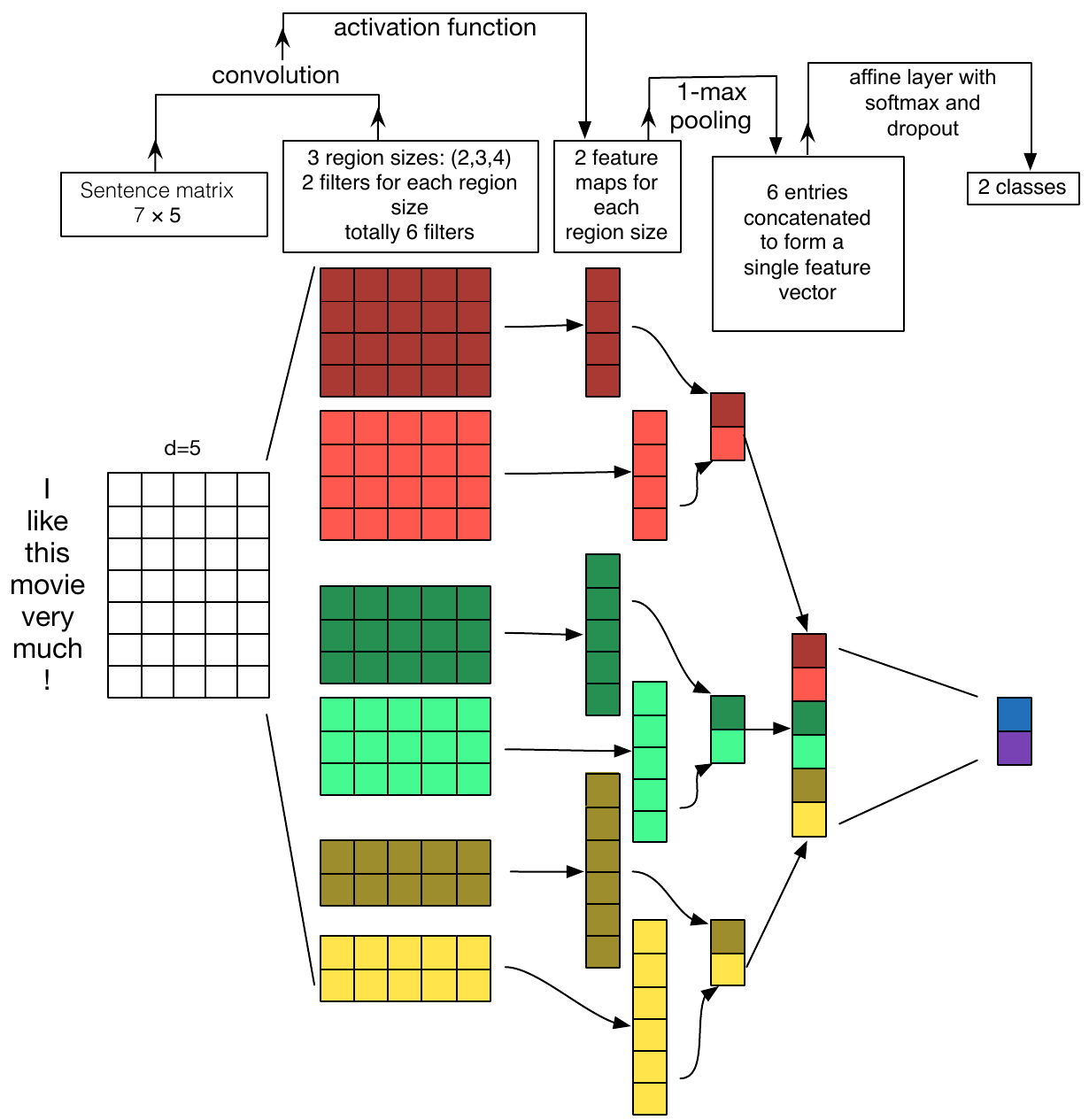}
\captionsetup{size=footnotesize}
\caption{\label{fig:cnn} CNN architecture for (short) document classification, taken from Zhang and Wallace (2015) \cite{Zhang2015practicioner}. $s=7$, $d=5$. 3 regions of respective sizes $h=\big\{2,3,4\big\}$ are considered, with associated output vectors of resp. lengths $s-h+1 =  \big\{6,5,4\big\}$ for each filter (produced after convolution, not shown). There are 2 filters per region size. For the three region sizes, the filters are resp. associated with feature maps of lengths $\big\{6,5,4\big\}$ (the output vectors after elementwise application of $f$ and addition of bias). $1$-max pooling is used.}
\end{figure}

\noindent For short sentence classification, best region sizes are generally found between 1 and 10, and in practice, $n_f$ filters (with $n_f \in [100,600]$) are applied to each region to give the model the ability to learn different, complementary features for each region \cite{Zhang2015practicioner}. Since each filter generates a feature map, each region is thus embedded into an $n_f$-dimensional space. Moreover, using regions of varying size around the optimal one improves performance \cite{Zhang2015practicioner}. In that case, different parallel branches are created (one for each region size), and the outputs are concatenated after pooling, as shown in Fig. \ref{fig:cnn}. Performance and cost increase with $n_f$ up to a certain point, after which the model starts overfitting. \\

\subsubsection{Pooling layer} The exact positions of the features in the input document do not matter. What matters is only whether certain features are present or absent. For instance, to classify a review as positive, whether ``best movie ever'' appears at the beginning or at the end of the document is not important. To inject such robustness into the model, \textit{global $k$-max pooling}\footnote{pooling may also be applied locally over small regions, but for short text classification, global pooling works better \cite{Zhang2015practicioner}.} is employed. This approach extracts the $k$ greatest values from each feature map and concatenates them, thus forming a final vector whose size always remains constant during training. For short sentence classification, \cite{Zhang2015practicioner} found that $k=1$ was by far superior to higher-order strategies. They also reported that using the maximum was much better than using the average, which makes sense, since we're only interested in extracting the most salient feature from each feature map.\\

\subsubsection{Document encoding} As shown in Fig. \ref{fig:cnn}, looking at things from a high level, the CNN architecture connects each filtered version of the input to a single neuron in a final feature vector. This vector can be seen as an embedding, or \textit{encoding}, of the input document. It is the main contribution of the model, the thing we're interested in. The rest of the architecture just depends on the task.

\subsubsection{Softmax layer}
Since the goal here is to classify documents, a softmax function is applied to the document encoding to output class probabilities. However, different tasks would call for different architectures: determining whether two sentences are paraphrases, for instance, would require two CNN encoders sharing weights, with a final energy function and a contrastive loss (\`a la Siamese \cite{chopra}); for translation or summarization, we could use a LSTM language model decoder conditioned on the CNN encoding of the input document  (\`a la \texttt{seq-to-seq} \cite{seq-to-seq}), etc.

Going back to our classification setting, the softmax transforms a vector $x \in \mathbb{R}^{K}$ into a vector of positive floats that sum to one, i.e., into a \textit{probability distribution} over the classes to be predicted:

\begin{equation}
\mathrm{softmax}(x_{i}) = \frac{e^{x_{i}}}{\sum_{j=1}^{K}e^{x_{j}}}
\end{equation}

\noindent In the binary classification case, instead of having a final output layer of two neurons with a softmax, where each neuron represents one of the two classes, we can have an output layer with only one neuron and a sigmoid function ($\sigma(x)=\frac{1}{1+e^{-x}}$). In that case, the neuron outputs the probability of belonging to one of the two classes, and decision regarding the class to predict is made based on whether $\sigma(x)$ is greater or smaller than 0.5 (assuming equal priors). These two approaches are equivalent. Indeed, $\frac{1}{1+e^{-x}}=\frac{e^{x}}{e^{x}+e^{0}}$. So, the one-neuron sigmoid layer can be viewed as a two-neuron softmax layer where one of the neurons never activates and has its output always equal to zero.

\subsection{Number of parameters}
The total number of trainable parameters for our CNN is the sum of the following terms:

\begin{itemize}
\item \textbf{word embedding matrix} (only if non-static mode): $(V+1) \times d$, where $V$ is the size of the vocabulary. We add one row for the zero-padding vector.
\item \textbf{convolution layer}: $h \times d \times n_f + n_f$ (the number of entries in each filter by the number of filters, plus the biases).
\item \textbf{softmax layer}: $n_f \times 1 + 1$ (fully connected layer with an output dimension of 1 and one bias).

\end{itemize}

\subsection{Visualizing and understanding inner representations and predictions}
\subsubsection{Document embeddings}
\begin{figure}[h]
\centering
\begin{minipage}{0.49\textwidth}
\centering
\includegraphics[width=0.99\textwidth]{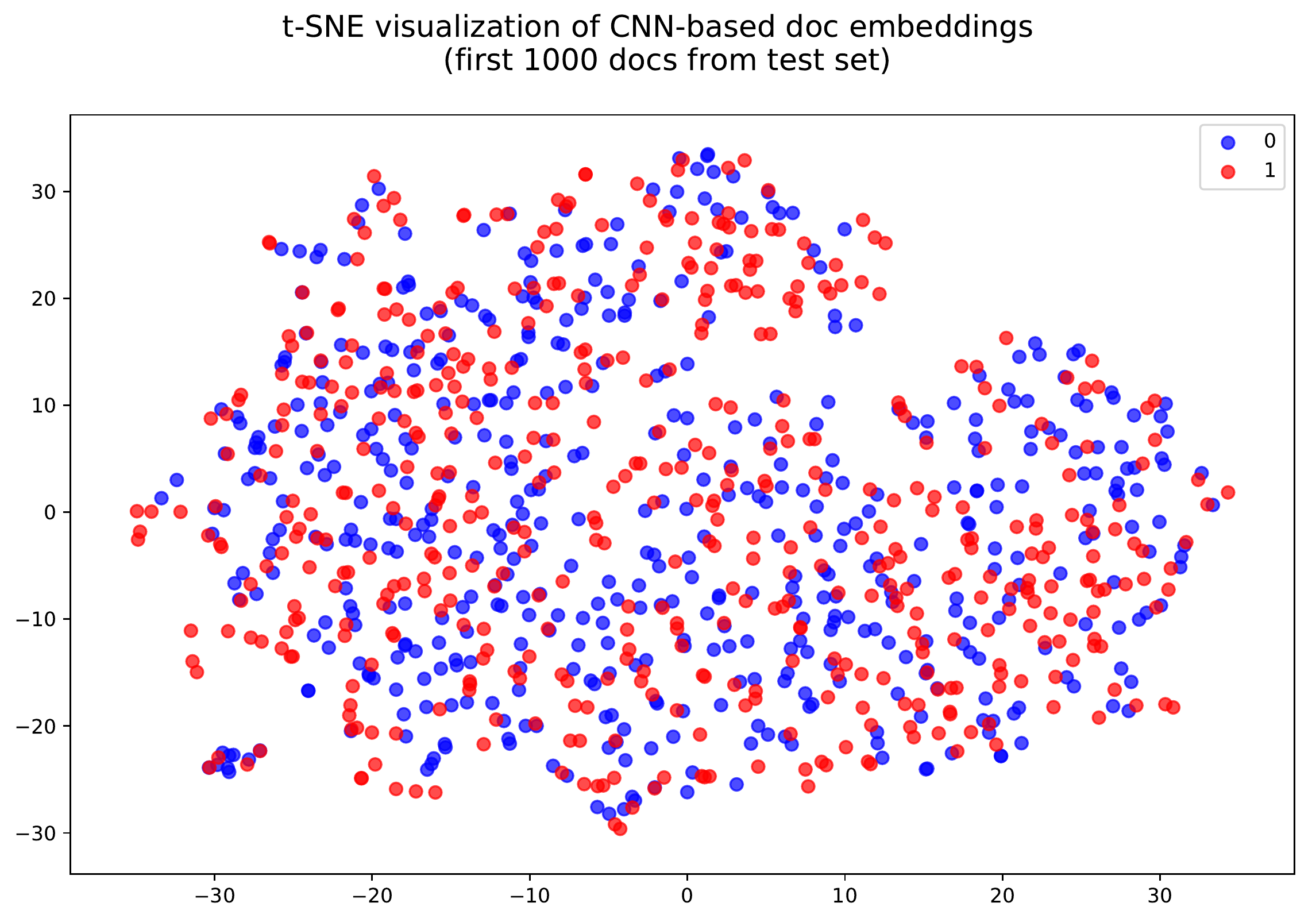}
\captionsetup{size=footnotesize}
\caption{\label{fig:doc_emb_init} Doc embeddings before training.}
\end{minipage}\hfill
\begin{minipage}{0.49\textwidth}
\centering 
\includegraphics[width=0.99\textwidth]{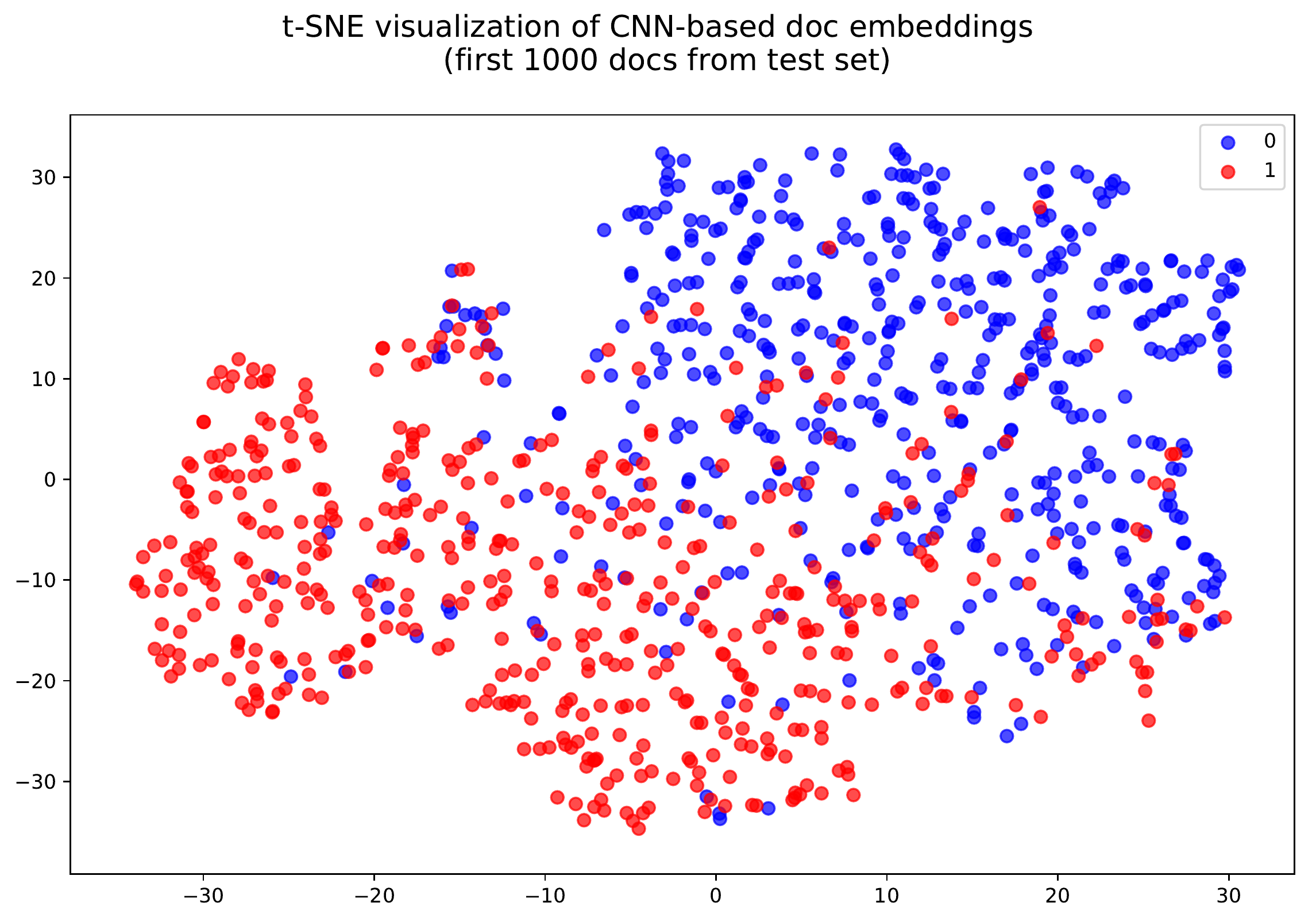}
\captionsetup{size=footnotesize}
\caption{\label{fig:doc_emb} Doc embeddings after 2 epochs.}
\end{minipage}
\end{figure}

A fast and easy way to verify that our model is learning effectively is to check whether its internal document representations make sense. Recall that the feature vector which is fed to the softmax layer can be seen as an $n_f$-dimensional encoding of the input document. By collecting the intermediate output of the model at this precise level in the architecture for a subset of documents, and projecting the vectors to a low-dimensional map, we can thus visualize whether there is any correlation between the embeddings and the labels. Fig.s \ref{fig:doc_emb_init} and \ref{fig:doc_emb} prove that indeed, our model is learning meaningful representations of documents.

\subsubsection{Predictive regions identification}
This approach is presented in section 3.6 (Tables 5 \& 6) of \cite{JohnsonZhang}. Recall that before we lose positional information by applying pooling, each of the $n_f$ filters of size $h$ is associated with a vector of size $\nicefrac{(s−h)}{\mathrm{stride}}+1$ (a feature map) whose entries represent the output of the convolution of the filter with the corresponding receptive field in the input, after application of the nonlinearity and addition of the bias. Therefore, each receptive field is embedded into an $n_f$-dimensional space. 
Thus, after training, we can identify the regions of a given document that are the most predictive of its category by inspecting the intermediate output of the model corresponding to the receptive field embeddings (right before the pooling layer), and by finding the regions that have the highest norms. For instance, some of the most predictive regions for negative IMDB reviews are: ``worst movie ever'', ``don't waste your money'', ``poorly written and acted'', ``awful picture quality''. Conversely, some regions very indicative of positivity are: ``amazing soundtrack'', ``visually beautiful'', ``cool journey'', ``ending quite satisfying''... \\

\subsubsection{Saliency maps}
Another way to understand how the model is issuing its predictions was described by \cite{simonyan2013deep} and applied to NLP by \cite{lin2015visualizing}. The idea is to rank the elements of the input document $A \in \mathbb{R}^{s \times d}$ based on their influence on the prediction. An approximation can be given by the magnitudes of the first-order partial derivatives of the output of the model $\mathrm{CNN}: A \mapsto \mathrm{CNN}(A)$ with respect to each row $a$ of $A$:

\begin{equation}
\mathrm{saliency}(a)=\abs{\frac{\partial(\mathrm{CNN})}{\partial a}\mid_a}
\end{equation}

\begin{figure}[h]
\centering
\includegraphics[width=0.68\textwidth]{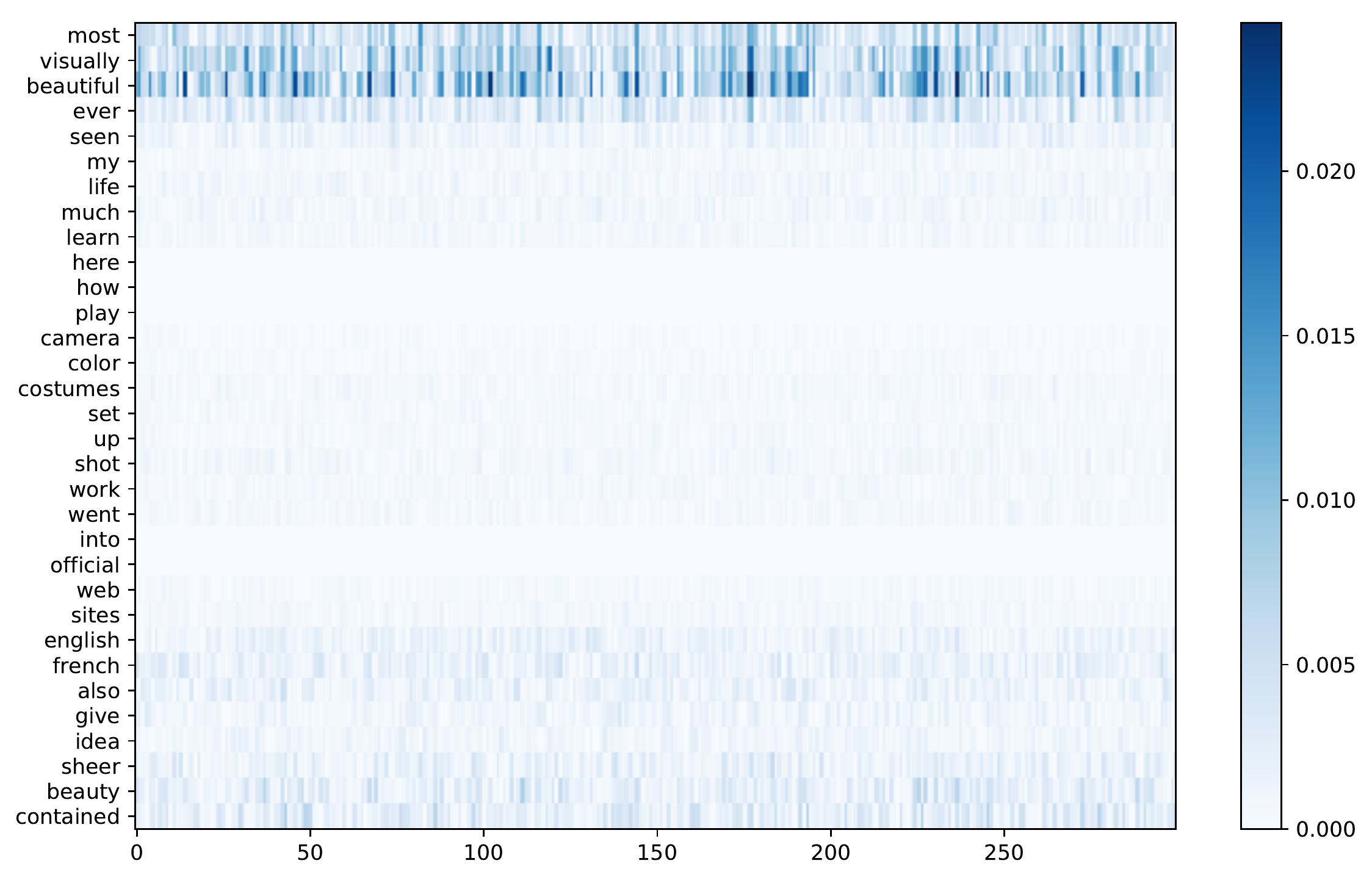}
\captionsetup{size=footnotesize}
\caption{\label{fig:saliency_pos} Saliency map for document 1 of the IMDB test set (true label: positive)}
\end{figure}

\begin{figure}[H]
\centering
\includegraphics[width=0.68\textwidth]{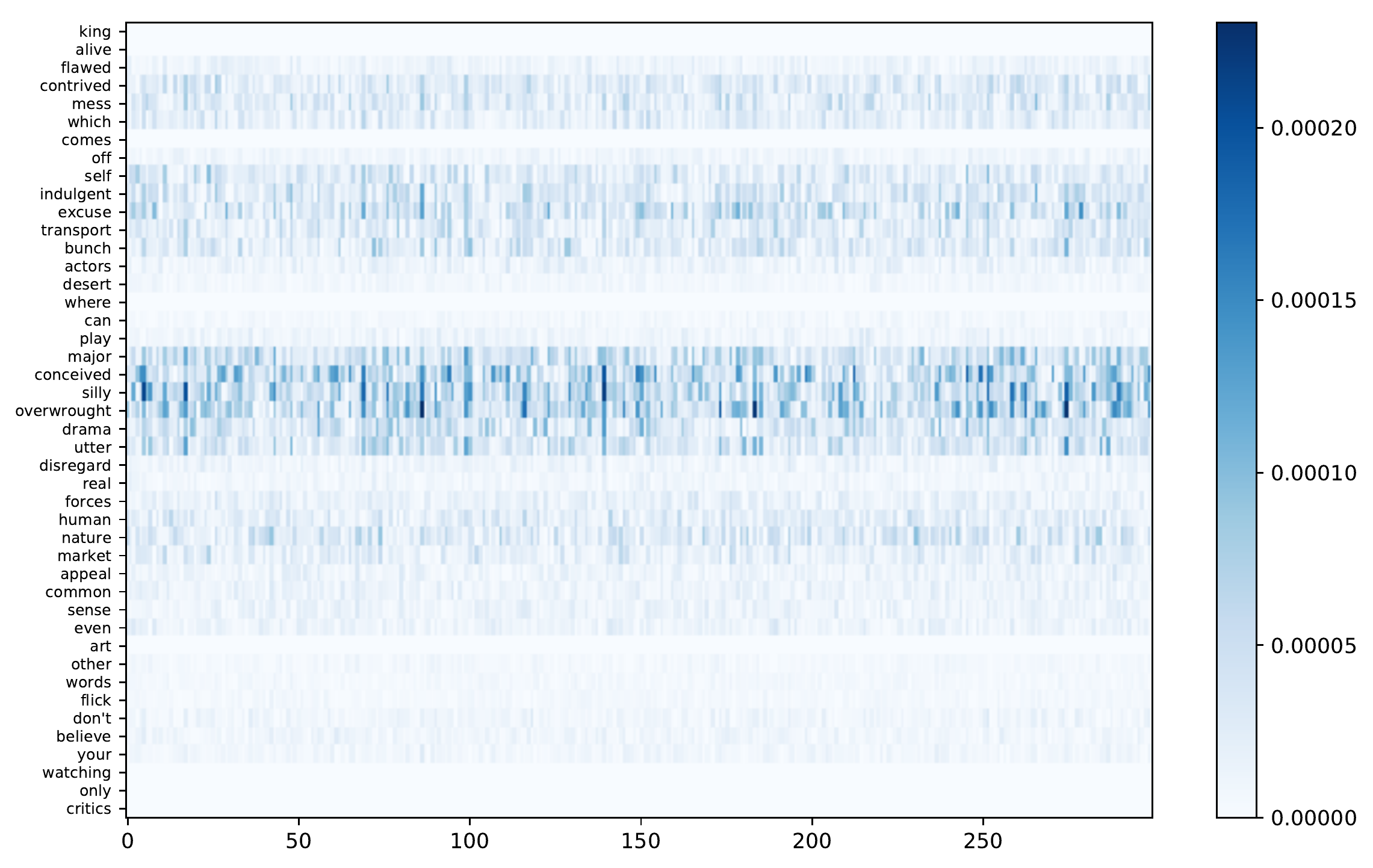}
\captionsetup{size=footnotesize}
\caption{\label{fig:saliency_neg} Saliency map for document 15 of the IMDB test set (true label: negative)}
\end{figure}

\noindent The interpretation is that we identify which words in $A$ need to be \textit{changed the least to change the class score the most}. The derivatives can be obtained by performing a single back-propagation pass (based on the prediction, not the loss like during training). Fig.s \ref{fig:saliency_pos} and \ref{fig:saliency_neg} show saliency map examples for negative and positive reviews, respectively.

\section{Recurrent Neural Networks (RNNs)}
We first present the overall RNN framework, and then two types of units widely used in practice: the LSTM and the GRU. A good review of RNNs, LSTMs and their applications can be found in \cite{lipton}.

\subsection{RNN framework}
\begin{figure}[H]
\centering
\includegraphics[width=0.42\textwidth]{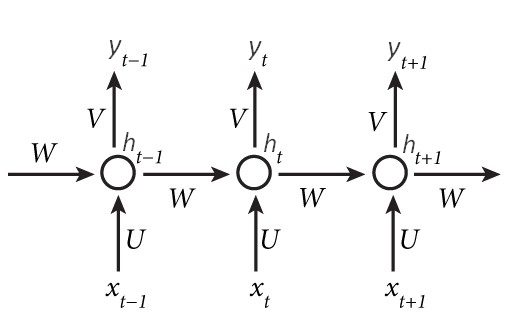}
\captionsetup{size=small}
\caption{\label{fig:rnn} 3 steps of an unrolled RNN (\textit{adapted from \href{http://www.wildml.com/2015/09/recurrent-neural-networks-tutorial-part-2-implementing-a-language-model-rnn-with-python-numpy-and-theano/}{Denny Britz' blog}}. Each circle represents a RNN unit (see equations \ref{eq:rnn_hidden} \& \ref{eq:rnn_output}).}
\end{figure}

While CNNs are naturally good at dealing with grids, RNNs were specifically developed to be used with \textit{sequences} \cite{elman1990finding}. Some examples include time series, or, in NLP, words (sequences of characters) or sentences (sequences of words). CNNs do allow to capture some order information, but it is limited to \textit{local} patterns, and long-range dependencies are ignored \cite{goldberg2016primer}. As shown in Fig. \ref{fig:rnn}, a RNN can be viewed as a chain of simple neural layers that \textit{share} the same parameters.

\begin{figure}[H]
\centering
\includegraphics[width=0.52\textwidth]{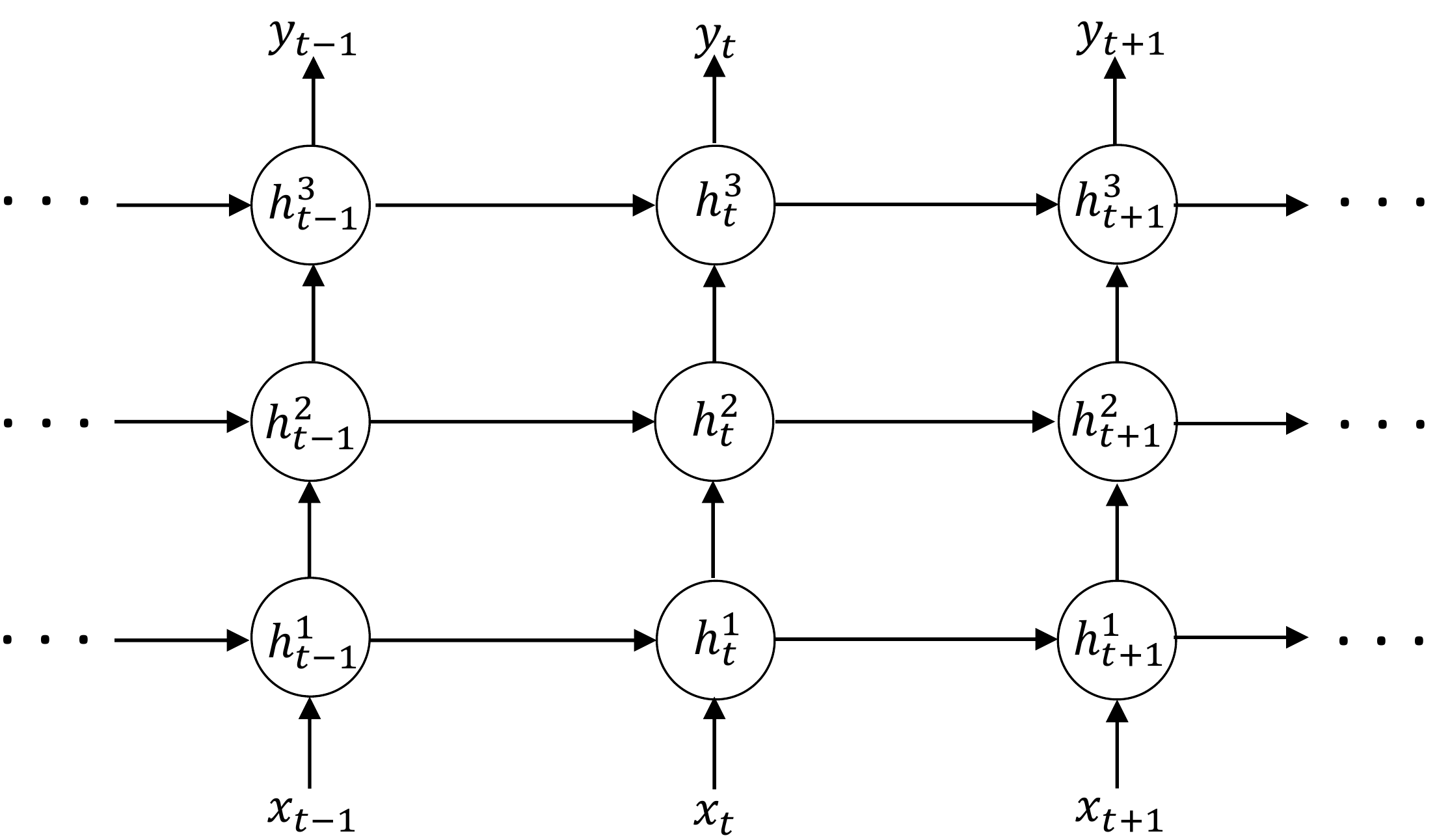}
\captionsetup{size=small}
\caption{\label{fig:hier_rnn} 3 steps of an unrolled deep RNN. Each circle represents a  RNN unit. The hidden state of each unit in the inner layers (1 \& 2) serves as input to the corresponding unit in the layer above.}
\end{figure}

\noindent From a high level, a RNN is fed an ordered list of input vectors $\big\{x_{1},...,x_{T}\big\}$ as well as an initial hidden state $h_{0}$ initialized to all zeros, and returns an ordered list of hidden states $\big\{h_{1},...,h_{T}\big\}$, as well as an ordered list of output vectors $\big\{y_{1},...,y_{T}\big\}$. The output vectors may serve as input for other RNN units, when considering deep architectures (multiple RNN layers stacked vertically, as shown in Fig. \ref{fig:hier_rnn}). The hidden states correspond more or less to the ``short-term'' memory of the network. Note that each training example is a full $\big\{x_{1},...,x_{T}\big\}$ sequence of its own, and may be associated with a label depending on the task. E.g., for short document classification, the sequences would be associated with a label, whereas for language modeling, we would just parse all sequences, repeatedly predicting the next words.

\noindent At any step $t$ in the sequence, the hidden state $h_{t}$ is defined in terms of the previous hidden state $h_{t-1}$ and the current input vector $x_{t}$ in the following \textit{recursive} way:

\begin{equation}\label{eq:rnn_hidden}
h_{t} = f(Ux_{t} + Wh_{t-1} + b)
\end{equation}

\noindent Where $f$ is a nonlinearity such as \texttt{tanh} (applied elementwise), $x_{t} \in \mathbb{R}^{d_{in}}$, $U\in\mathbb{R}^{H \times d_{in}}$ and  $W\in\mathbb{R}^{H \times H}$ are parameter matrices shared by all time steps, and $h_{t}$, $h_{t-1}$ and $b$ belong to $\mathbb{R}^{H}$. $d_{in}$ can be the size of the vocabulary, if one-hot vectors are passed as input, or the dimensionality of the embedding space, when working with shared features. $H$ is the dimension of the hidden layer. Usually, $H \sim 100$. The larger this layer, the greater the capacity of the memory, with an increase in computational cost.

The output vector $y_{t} \in \mathbb{R}^{d_{out}}$ transforms the current hidden state $h_{t} \in \mathbb{R}^{H}$ in a way that depends on the final task. For classification, it is computed as:

\begin{equation}\label{eq:rnn_output}
y_{t} = \mathrm{softmax}(Vh_{t})
\end{equation}

\noindent Where $V\in\mathbb{R}^{d_{out} \times H}$ is a parameter matrix shared across all time steps. $d_{out}$ depends on the number of categories. E.g., for 3-class document classification, $d_{out}=3$, for a word-level language model, $d_{out}=|V|$. 

Note that when stacking multiple RNN layers vertically (deep RNN architecture), the hidden states of the units below are directly connected to the units above, i.e., $x_{t_{above}} = y_{t_{below}}$ and $y_{t_{below}} = h_{t_{below}}$. The output layer (Eq. \ref{eq:rnn_output}) lies on top of the stack.

\subsubsection{Language modeling}
Language modeling is a special case of classification where the model is trained to predict the next word or character in the sentence. At each time step $t$, the output vector gives the probability distribution of $x_t$ over all the words/characters in the vocabulary, conditioned on the previous words/characters in the sequence, that is, $P[x_t|x_{t-1},...,x_1]$. At test time, the probability of a full sequence $\small\{x_{1},...,x_{T}\small\}$ is given by the product of all conditional probabilities as:

\begin{equation}
P\big[\small\{x_{1},...,x_{T}\small\}\big] = P[x_1] \times \prod_{t=2}^{T} P[x_t|x_{t-1},...,x_1] 
\end{equation}

\noindent The language model can also be used to generate text of arbitrary size by repeatedly sampling characters for the desired number of time steps (for character-level granularity) or until the special end-of-sentence token is selected\footnote{see \cite{graves_2013} and \url{http://karpathy.github.io/2015/05/21/rnn-effectiveness/}} (for word-level granularity).

For a character-level language model for instance, $T$ can easily exceed 20 or 25. This greatly amplifies the adverse effects of the well-known \textit{vanishing} and \textit{exploding} gradients problem, which prevents long-range dependencies from being learned\footnote{\href{foo}{wildml.com/2015/10/recurrent-neural-networks-tutorial-part-3-backpropagation-through-time-and-vanishing-gradients/}}. Note that this issue can also be experienced with feed-forward neural networks, such as the Multi-Layer Perceptron, but it just gets worse with RNN due to their inherent tendency to be deep.

\subsection{LSTM unit}
In practice, whenever people use RNNs, they use the LSTM or the GRU unit (see next subsection), as these cells are engineered in a way that allows them to escape vanishing/exploding gradients and keep track of information over longer time periods \cite{1997lstm}.

\begin{figure}[H]
\centering
\includegraphics[width=0.55\textwidth]{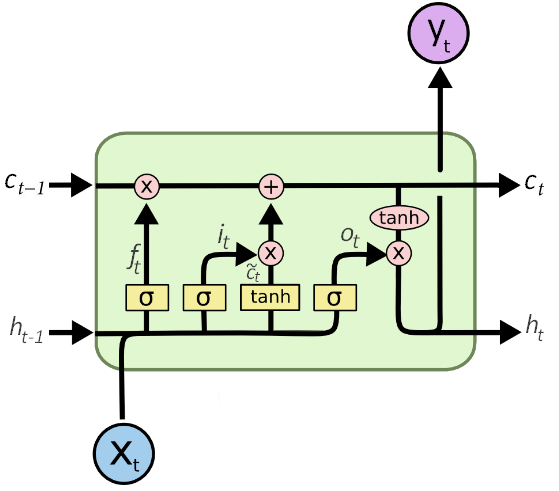}
\captionsetup{size=small}
\caption{\label{fig:lstm} The LSTM unit. Adapted from \textit{\href{http://colah.github.io/posts/2015-08-Understanding-LSTMs/}{Chris Colah's blog}}.}
\end{figure}

\noindent As shown in Fig. \ref{fig:lstm}, the two things that change in the LSTM unit compared to the basic RNN unit are (1) the presence of a \textit{cell state} ($c_{t})$, which serves as an explicit memory, and (2) how hidden states are computed. With vanilla RNNs, the hidden state is computed with a single layer as $h_{t} = \mathrm{tanh}(Ux_{t} + Wh_{t-1} + b)$ (see eq. \ref{eq:rnn_hidden}). With the LSTM unit however, the hidden state is computed by four interacting layers that give the network the ability to remember or forget specific information about the preceding elements in the sequence.

\subsubsection{Inner layers}
The four layers are:

\begin{enumerate}
\item forget gate layer: $f_{t} = \sigma \big(U_{f}x_{t} + W_{f}h_{t-1} + b_f\big)$
\item input gate layer: $i_{t} = \sigma \big(U_{i}x_{t} + W_{i}h_{t-1} + b_i\big)$
\item candidate values computation layer: $\tilde{c}_{t} = \mathrm{tanh} \big(U_{c}x_{t} + W_{c}h_{t-1} + b_c\big)$
\item output gate layer: $o_{t} = \sigma \big(U_{o}x_{t} + W_{o}h_{t-1} + b_o\big)$
\end{enumerate}
Thanks to the elementwise application of the sigmoid function ($\sigma$), the forget, input, and output \textit{gate} layers (1, 2, and 4 above) generate vectors whose entries are all comprised between 0 and 1, and either close to 0 or close to 1. When one of these layers is multiplied with another vector, it thus acts as a filter that only selects a certain proportion of that vector. This is precisely why those layers are called gates. The two extreme cases are when all entries are equal to 1 -the full vector passes- or to 0 -nothing passes. Note that the 3 forget, input, and output gates are computed in the exact same way, only the parameters vary. The parameters are however shared across all time steps.

\subsubsection{Forgetting/learning}
By taking into account the new training example $x_{t}$ and the current hidden state $h_{t-1}$, the forget gate layer $f_{t}$ determines how much of the previous cell state $c_{t-1}$ should be forgotten (what fraction of the memory should be freed up), while from the same input, the input gate layer $i_{t}$ decides how much of the candidate values $\tilde{c}_{t}$ should be written to the memory, or in other words, how much of the new information should be learned. Combining the output of the two filters \textit{updates} the cell state:

\begin{equation}\label{eq:ct}
c_{t} = f_{t} \circ {c}_{t-1} + i_{t} \circ \tilde{c}_{t}
\end{equation}

\noindent Where $\circ$ denotes elementwise multiplication (Haddamard product). This way, important information is not overwritten by the new inputs but is able to be kept alongside them for long periods of time. Finally, the activation $h_t$ is computed from the updated memory, modulated by the output gate layer $o_{t}$:

\begin{equation}\label{eq:out}
h_{t} = \mathrm{tanh}\big(c_{t}\big) \circ o_{t}
\end{equation}

\noindent The output gate allows the unit to only activate when the in-memory information is found to be relevant for the current time step. Finally, as before with the simple RNN, the output vector is computed as a function of the new hidden state:

\begin{equation}\label{eq:yt}
y_{t} = \mathrm{softmax}(Vh_{t})
\end{equation}

\subsubsection{Vanilla RNN analogy} 
If we decide to forget everything about the previous state (all elements of $f_{t}$ are null), to learn all of the new information (all elements of $i_{t}$ are equal to 1), and to memorize the entire cell state to pass to the next time step (all elements of $o_{t}$ are equal to 1), we have $c_{t} = \tilde{c}_{t} = \mathrm{tanh} \big(U_{c}x_{t} + W_{c}h_{t-1} + b_c\big)$, and thus we go back to a vanilla RNN unit, the only difference being an additional \texttt{tanh}, as we end up with $h_{t} = \mathrm{tanh}\big(\mathrm{tanh} \big(U_{c}x_{t} + W_{c}h_{t-1} + b_c\big)\big)$ instead of $h_{t} = \mathrm{tanh} \big(U_{c}x_{t} + W_{c}h_{t-1} + b_c\big)$ like in the classical RNN case.

\subsection{Gated Recurrent Unit (GRU)}
\begin{figure}[H]
\centering
\includegraphics[width=0.5\textwidth]{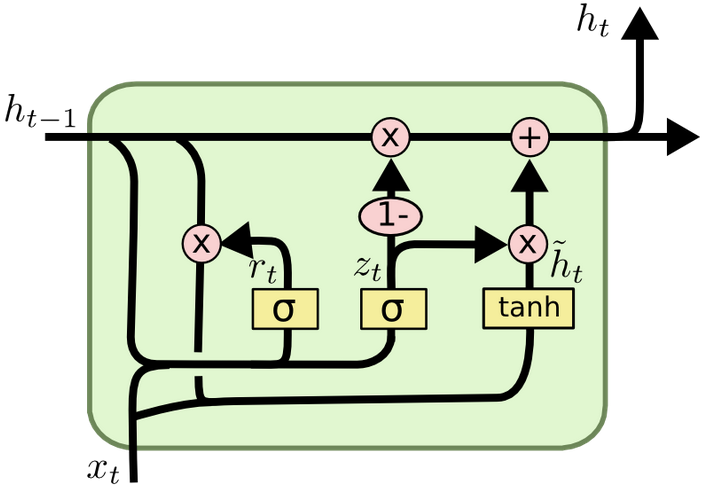}
\captionsetup{size=small}
\caption{GRU unit. Taken from \textit{\href{http://colah.github.io/posts/2015-08-Understanding-LSTMs/}{Chris Colah's blog}}. \label{fig:gru}}
\end{figure}

As shown in Fig. \ref{fig:gru}, the GRU unit \cite{gru} is a simplified LSTM unit with only two gates (reset and update), and where there is no explicit memory $c_t$.

\begin{enumerate}
\item reset gate layer: $r_{t} = \sigma \big(U_{r}x_{t} + W_{r}h_{t-1} + b_r\big)$
\item update gate layer: $z_{t} = \sigma \big(U_{z}x_{t} + W_{z}h_{t-1} + b_z\big)$
\end{enumerate}

\noindent The candidate hidden state is computed as:

\begin{equation}\label{eq:gru_candidate}
\tilde{h}_{t} = \mathrm{tanh} \big(U_{h}x_{t} + W_{h} (r_t \circ h_{t-1}) + b_h\big)\footnote{It should be noted that the original formulation of \cite{gru} uses $r_t \circ (W_{h} h_{t-1})$. Here, we adopt the formulation of \cite{gru_emp}, $W_{h} (r_t \circ h_{t-1})$. According to \cite{gru_emp}, the two formulations perform equivalently.}
\end{equation}

\noindent When all elements of the reset gate approach zero, information from the previous time steps (stored in $h_{t-1}$) is discarded, and the candidate hidden state is thus only based on the current input $x_t$. The new hidden state is finally obtained in a way similar to that of the LSTM cell state, by linearly interpolating between the previous hidden state and the candidate one:

\begin{equation}\label{eq:gru_hidden}
h_{t} = z_t \circ {h}_{t-1} + (1-z_t) \circ \tilde{h}_{t}
\end{equation}

\noindent the only difference is that this time, the update gate $z_t$ serves as the forget gate and determines the fraction of information from the previous hidden state that should be forgotten, and the input gate is \textit{coupled} on the forget gate.

\subsection{RNN vs LSTM vs GRU}
The basic RNN unit exposes its full hidden state at every time step (see Eq. \ref{eq:rnn_hidden}), so as time goes by, the impact of older inputs is quickly replaced by that of the more recent ones. The RNN is therefore not able to remember important features for more than a few steps. Indeed, we have shown previously that a RNN is analogous to a LSTM where for all $t$, $f_t=\vec{0}$, $i_t=\vec{1}$, and $o_t=\vec{1}$ (we forget everything about the past and learn everything about the present).\\

\noindent On the other hand, thanks to the use of an explicit memory (the cell) and a gating mechanism, the LSTM unit is able to control which fraction of information from the past should be kept in memory (forget gate $f_t$), which fraction of information from the current input should be written to memory (input gate $i_t$), and how much of the memory should be exposed to the next time steps and to the units in the higher layers (output gate $o_t$).\\

\noindent The GRU also features a gating mechanism, but has no explicit memory (no cell state). As a result, the gating mechanism of the GRU is simpler, without output gate: the linear interpolation between the old and the new information is directly injected into the new hidden state without filtering (see Eq. \ref{eq:gru_hidden}). Another difference is that when computing the candidate values, the GRU, via its reset gate $r_t$, modulates the flow of information coming from the previous activation $h_{t-1}$ (see Eq. \ref{eq:gru_candidate}), while in the LSTM unit, $\tilde{c}_{t}$ is based on the raw $h_{t-1}$. Last but not least, in the GRU, the balance between the old and the new information is only made by the update gate $z_t$ (see Eq. \ref{eq:gru_hidden}), whereas the LSTM unit has two \textit{independent} forget and input gates. \\

\noindent While both the LSTM and GRU units are clearly superior to the basic RNN unit \cite{gru_emp}, there is no evidence about which one is best \cite{lstm_odyssey,gru_emp}. However, since the GRU is simpler, it is easier to implement, more efficient, and has less parameters so it requires less training data.

\section{Attention}
The attention mechanism \cite{att_bahdanau} was developed in the context of encoder-decoder architectures for Neural Machine Translation (NMT) \cite{gru,seq-to-seq}, and rapidly applied to naturally related tasks such as image captioning (translating an image to a sentence) \cite{show_attend_tell}, and summarization (translating to a more compact language) \cite{att_summ}. From a high-level, by allowing the decoder to shop for what it needs over multiple vectors, attention relieves the encoder from the burden of having to embed the input into a single fixed-length vector, and thus allows to keep much more information \cite{att_bahdanau}.

Today, attention is ubiquitous in deep learning models, and is not used only in encoder-decoder contexts. Notably, attention devices have been proposed for encoders only, to solve tasks such as document classification \cite{han} or representation learning \cite{infersent}. Such mechanisms are qualified as \textit{self} or \textit{inner} attention. 

In what follows, we will start by presenting attention in the original context of encoder-decoder for NMT, using the general framework introduced by \cite{att_luong}, and then introduce self-attention.

\subsection{Encoder-decoder attention}
\subsubsection{Encoder-decoder overview}
\begin{figure}[H]
\centering
\includegraphics[width=0.45\textwidth]{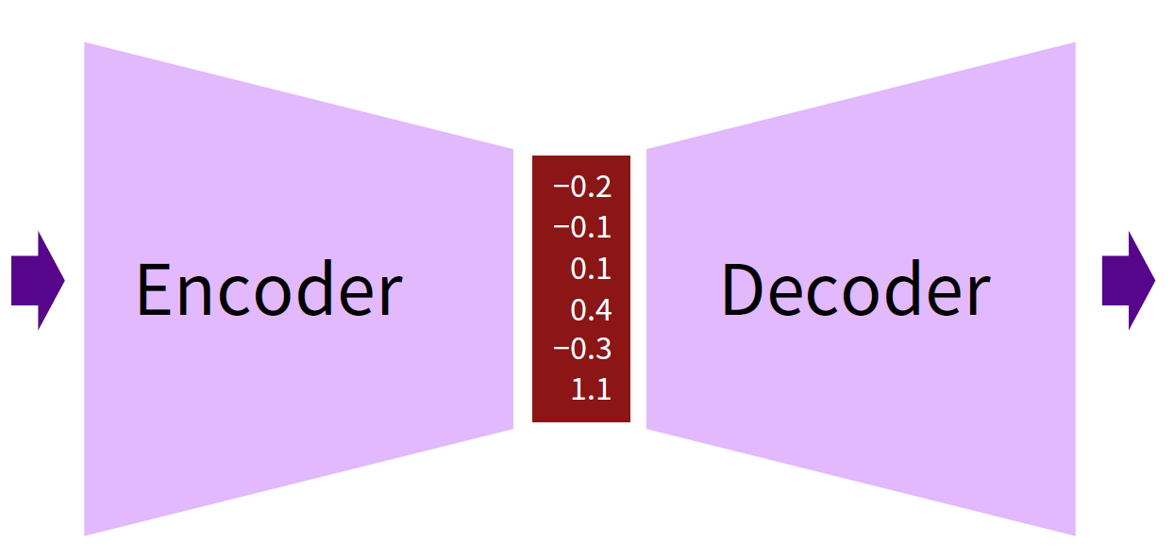}
\captionsetup{size=small}
\caption{Overview of the encoder-decoder architecture. Taken from \url{https://sites.google.com/site/acl16nmt/home} \label{fig:enc_dec_overview}}
\end{figure}

From a very high level, as shown in Fig. \ref{fig:enc_dec_overview}, the encoder embeds the input into a vector, and the decoder generates some output from this vector.

\noindent In Neural Machine Translation (NMT), the input and the output are sequences of words, respectively $x = \big(x_1, \dots ,x_{T_x}\big)$ and $y = \big(y_1, \dots ,y_{T_y}\big)$. $x$ and $y$ are usually referred to as the \textit{source} and \textit{target} sentences. When both the input and the output are sequences, encoder-decoder architectures are sometimes called sequence-to-sequence (seq2seq) \cite{seq-to-seq}. Thanks to the fact that encoder-decoder architectures are differentiable everywhere, their parameters $\theta$ can be simultaneously optimized with maximum likelihood  estimation (MLE) over a parallel corpus. This way of training is called \textit{end-to-end}.

\begin{equation}
\mathrm{argmax}_{\theta}\Bigg\{\sum_{(x,y) \in \mathrm{corpus}} \mathrm{log}~p(y|x;\theta) \Bigg\}
\end{equation}

\noindent Here, the function that we want to maximize is the log probability of a correct translation.

\subsubsection{Encoder}
The source sentence can be embedded by any model (e.g., CNN, fully connected). Usually for MT though, the encoder is a deep RNN. Bahdanau et al. \cite{att_bahdanau} originally used a \textit{bidirectional} deep RNN. Such a model is made of two deep unidirectional RNNs, with different parameters except the word embedding matrix. The first \textit{forward} RNN processes the source sentence from left to right, while the second \textit{backward} RNN processes it from right to left. The two sentence embeddings are concatenated at each time step $t$ to obtain the inner representation of the bidirectional RNN:

\begin{equation}\label{eq:bi_gru}
h_{t} = \big[ \vec{h_t}; \cev{h_t} \big]
\end{equation}

\noindent The bidirectional RNN takes into account the entire context when encoding the source words, not just the preceding words. As a result, $h_t$ is biased towards a small window centered on word $x_t$, while with a unidirectional RNN, $h_t$ is biased towards $x_t$ and the words immediately preceding it. Focusing on a small window around $x_t$ may be advantageous, but does not seem crucial. Indeed, Luong et al. \cite{att_luong} obtained state-of-the-art results with a usual unidirectional deep RNN encoder. In what follows, the hidden states of the encoder will be written $\bar{h}_t$. They are sometimes called \textit{annotations} in the literature. 

\subsubsection{Decoder}
While different models can be used as the encoder, in NMT the decoder is usually a unidirectional RNN because this model is naturally adapted to the sequential nature of the generation task, and is usually deep (stacking). The decoder generates each word of the target sentence one step at a time.\\

\noindent \textbf{Key idea}. Making the decoder use only the last annotation $h_{T_x}$ produced by the encoder to generate output forces the encoder to fit as much information as possible into $h_{T_x}$. Since $h_{T_x}$ is a single fixed-size vector, its capacity is limited, so  some information is lost. 
On the other hand, the attention mechanism allows the decoder to consider the entire sequence $\big(h_1, \dots, h_{T_x}\big)$ of annotations produced by the encoder at each step of the generation process. As a result, the encoder is able to keep much more information by distributing it among all its annotations, knowing that the decoder will be able to decide later on which vectors it should pay attention to. \\

\noindent More precisely, the target sentence $y=(y_1,\dots,y_{T_y})$ is generated one word $y_t$ at a time based on the distribution:
\begin{equation}
P\big[y_t|\{y_{1},...,y_{t-1}\},c_t\big] = \mathrm{softmax}\big(W_s\tilde{h}_t\big)
\end{equation}

\noindent where $\tilde{h}_t$, the \textit{attentional} hidden state, is computed as:

\begin{equation}
\tilde{h}_t = \mathrm{tanh}\big(W_c\big[c_t;h_t\big]\big)
\end{equation}

\begin{figure}[H]
\centering
\includegraphics[width=0.99\textwidth]{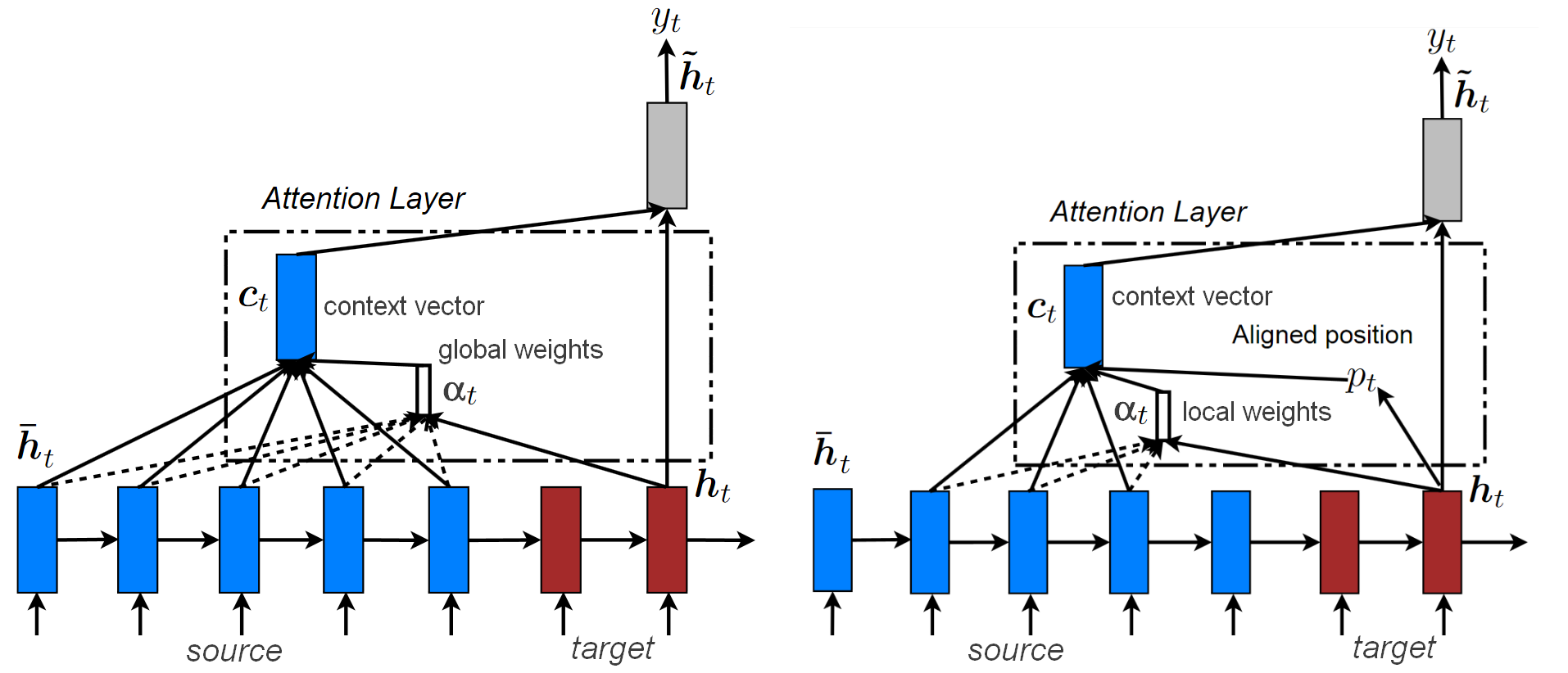}
\captionsetup{size=small}
\caption{Global (left) vs local attention (right). Adapted from \cite{att_luong}.\label{fig:global_vs_local_att}}
\end{figure}

\noindent $h_t$ is the hidden state of the decoder (hidden state of the top layer, when the decoder is a stacking RNN) and provides information about the previously generated target words $\{y_{1},...,y_{t-1}\}$, $c_t$ is the source context vector, and $\big[;\big]$ is concatenation. $W_s$ and $W_c$ are matrices of trainable parameters. Biases are not shown for simplicity. As shown in Fig. \ref{fig:global_vs_local_att}, the context vector $c_t$ can be computed in two ways: \textit{globally} and \textit{locally}. We describe each approach in the next two subsections.\\

\noindent \textbf{A note on beam search}. Trying all possible combinations of words in the vocabulary to find the target sentence with highest joint probability is intractable. But on the other hand, generating $y$ in a purely greedy way, i.e., by selecting the most likely word every time, is highly suboptimal. In practice, a certain number $K$ of candidate translations are explored with \textit{beam search}, a heuristic search algorithm \cite{beam}. Large values of $K$ generate better target sentences, but decrease decoding speed.

\subsubsection{Global attention}
Here, the context vector $c_t$ is computed as a weighted sum of the \textit{full list} of annotations $\bar{h}_i$ of the source sentence (i.e., the hidden states of the encoder). There are $T_x$ annotations. Each one is a vector of size the number of units in the hidden layer of the encoder. $c_t$ has same size as any annotation. The size of the alignment vector $\alpha_{t}$ is equal to the size $T_x$ of the source sentence, so it is \textit{variable}.

\begin{equation}
c_t = \sum_{i=1}^{T_x} \alpha_{t,i}\bar{h}_i
\end{equation}

\noindent The alignment vector $\alpha_{t}$ is computed by applying a softmax to the output of an \textit{alignment} operation ($\texttt{score()}$) between the current target hidden state $h_t$ and all source hidden states $\bar{h}_{i}$'s:

\begin{equation}\label{eq:weights_global}
\alpha_{t,i} = \frac{\mathrm{exp}\big(\mathrm{score}( h_t,\bar{h}_i)\big)}{\sum_{i'=1}^{T_x} \mathrm{exp}\big(\mathrm{score}( h_t,\bar{h}_{i'})\big)}
\end{equation}

\begin{figure}[H]
\centering
\includegraphics[width=0.95\textwidth]{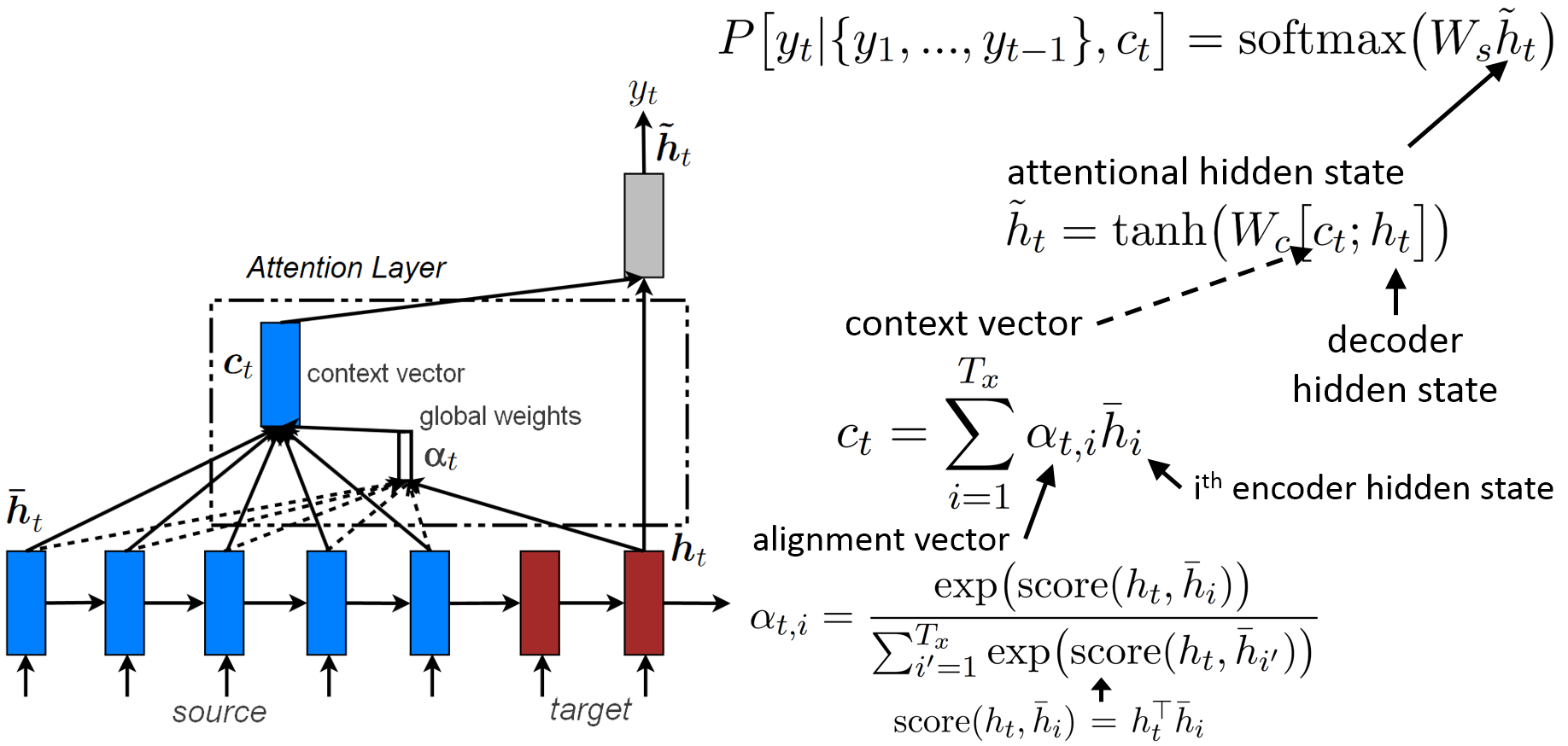}
\captionsetup{size=small}
\caption{Summary of the \textit{global attention} mechanism \cite{att_luong}.}\label{fig:global_summary}
\end{figure}

\noindent In other words, $\alpha_{t}$ is a probability distribution over all source hidden states (its coefficients are all between 0 and 1 and sum to 1), and indicates which words in the source sentence are the most likely to help in predicting the next word. \texttt{score()} can in theory be any comparison function. Luong et al. \cite{att_luong} experimented with the dot product ($\mathrm{score}(h_t,\bar{h}_i) = h_{t}^\top\bar{h}_i$), a more general formulation with a matrix of parameters ($\mathrm{score}(h_t,\bar{h}_i) = h_{t}^{\top}W_{\alpha}\bar{h}_i$), and a fully connected layer. They found that \textit{dot} works better for global attention while \textit{general} is superior for local attention. A summary of global attention is provided in Fig. \ref{fig:global_summary}.

\subsubsection{Local attention}
Considering all words in the source sentence to generate every single target word is expensive, and may not be necessary. To remediate this issue, Luong et al. \cite{att_luong} proposed to focus only on a small window of annotations of fixed size $2D + 1$:

\begin{equation}
c_t = \sum_{i=p_t-D}^{p_t+D} \alpha_{t,i}\bar{h}_i
\end{equation}

\noindent $D$ is prescribed by the user, and the position $p_t$ where to center the window is either set to $t$ (\textit{monotonic} alignment) or determined by a differentiable mechanism (\textit{predictive} alignment) based on information about the previously generated target words $\{y_{1},...,y_{t-1}\}$ stored in $h_t$:

\begin{equation}
p_t = T_x\cdot\sigma\big(v_{p}^{\top}\mathrm{tanh}(W_ph_t)\big)
\end{equation}

\noindent Where $T_x$ is the length of the source sentence, $\sigma$ is the sigmoid function, and $v_p$ and $W_p$ are trainable parameters. Alignment weights are computed like in the case of global attention (Eq. \ref{eq:weights_global}), with the addition of a Normal distribution term centered on $p_t$ and with standard deviation $\nicefrac{D}{2}$:

\begin{equation}\label{eq:weights_local}
\alpha_{t,i} = \frac{\mathrm{exp}\big(\mathrm{score}( h_t,\bar{h}_i)\big)}{\sum_{i'=p_t-D}^{p_t+D} \mathrm{exp}\big(\mathrm{score}( h_t,\bar{h}_{i'})\big)} \mathrm{exp}\Big(-\frac{(i-p_t)^2}{2(\nicefrac{D}{2})^2}\Big)
\end{equation}

\begin{figure}[H]
\centering
\includegraphics[width=0.95\textwidth]{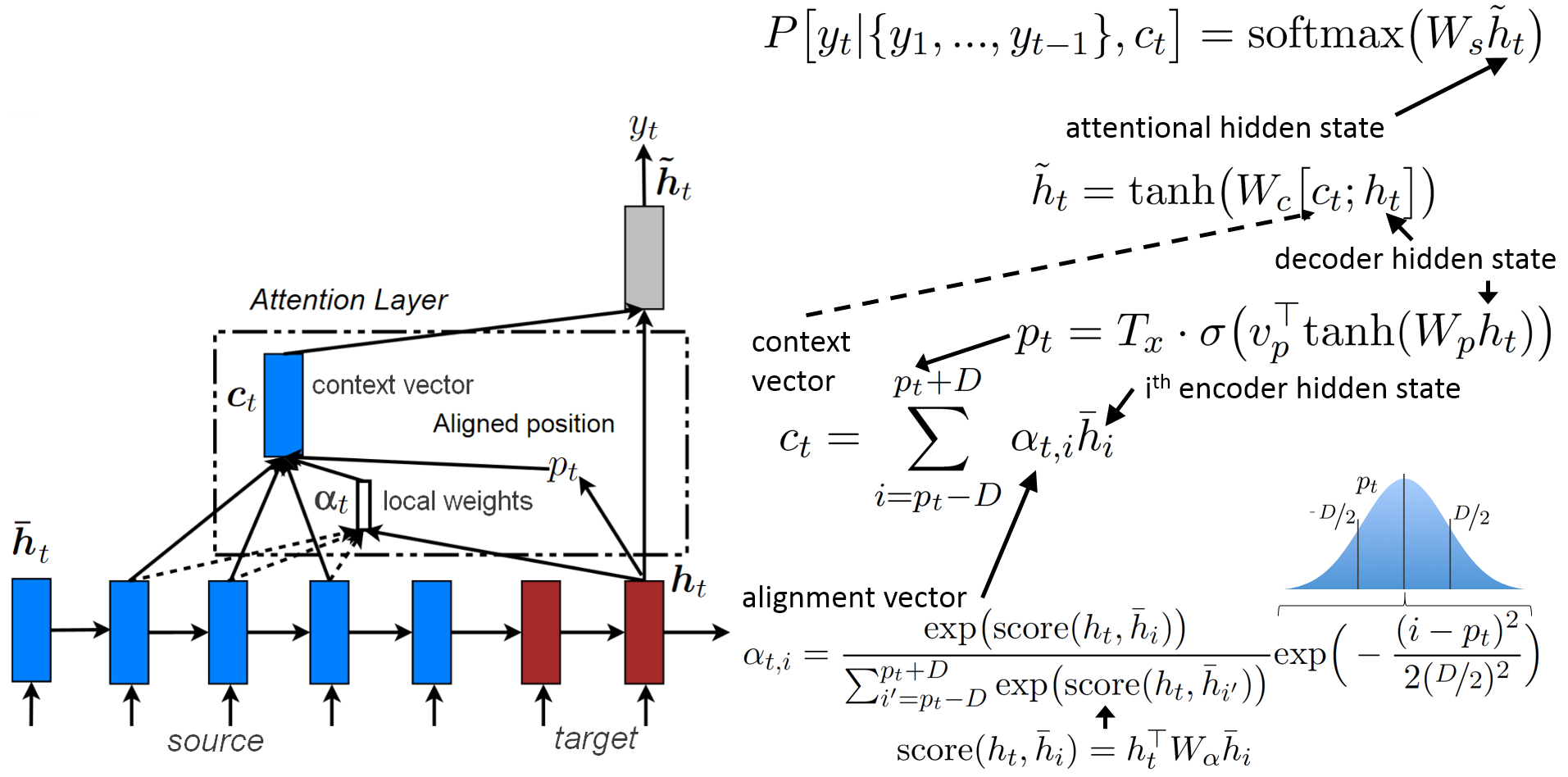}
\captionsetup{size=small}
\caption{Summary of the \textit{local attention with predictive alignment} mechanism \cite{att_luong}.}\label{fig:local_summary}
\end{figure}

\noindent Note that $p_t \in \mathbb{R} \cap \big[0,T_x\big]$ and $i \in \mathbb{N} \cap \big[p_t-D,p_t+D]$. The addition of the Gaussian term makes the alignment weights decay as $i$ moves away from the center of the window $p_t$, i.e., it gives more importance to the annotations near $p_t$. Also, unlike with global attention, the size of $\alpha_t$ is fixed and equal to $2D+1$, as only the annotations within the window are taken into account. Local attention can actually be viewed as global attention where alignment weights are multiplied by a truncated Normal distribution (i.e., that returns zero outside the window). A summary of local attention is provided in Fig. \ref{fig:local_summary}.

\subsection{Self-attention}
We are here in a simpler setting with a single RNN encoder taking as input a sequence $\big(x_1, \dots ,x_{T}\big)$ of length $T$. As usual, the RNN maps the input sequence to a sequence of annotations $\big(h_1, \dots, h_T\big)$. The goal is exactly the same as with attention in the encoder-decoder context: rather than considering the last annotation $h_T$ as a comprehensive summary of the entire sequence, which is prone to information loss, a new hidden representation is computed by taking into account the annotations at \textit{all} time steps. To this purpose, the self-attention or inner attention mechanism emerged in the literature in 2016/2017, with, e.g., \cite{han,self-attentive}. In what follows we use the formulation of \cite{han}. 

As shown in Eq. \ref{eq:self-att}, annotation $h_t$ is first passed to a dense layer. An alignment coefficient $\alpha_t$ is then derived by comparing the output $u_t$ of the dense layer with a trainable context vector $u$ (initialized randomly) and normalizing with a softmax. The attentional vector $s$ is finally obtained as a weighted sum of the annotations.


\begin{equation}\label{eq:self-att}
  \begin{split}
    u_t &= \mathrm{tanh}(Wh_t)\\
    \alpha_t &= \frac{\exp(\mathrm{score}(u_{t},u))}{\sum_{t'=1}^T \exp(\mathrm{score}(u_{t'},u))}\\
    s &= \sum_{t=1}^T \alpha_t h_t
  \end{split}
\end{equation}

\noindent $\mathrm{score}$ can in theory be any alignment function. A straightforward approach is to use $\mathrm{score}(u_{t},u) = u_{t}^\top u$. The context vector can be interpreted as a representation of the optimal word, on average. When faced with a new example, the model uses this knowledge to decide which word it should pay attention to. During training, through backpropagation, the model updates the context vector, i.e., it adjusts its internal representation of what the optimal word is.

\subsubsection{Difference with seq2seq attention}
The context vector in the definition of self-attention above has nothing to do with the context vector used in seq2seq attention! In seq2seq, the context vector $c_t$ is equal to the weighted sum $\sum_{i=1}^{T_x} \alpha_{t,i}\bar{h}_i$, and is used to compute the attentional hidden state as $\tilde{h}_t = \mathrm{tanh}\big(W_c\big[c_t;h_t\big]\big)$. In self-attention however, the context vector is simply used as a replacement for the hidden state of the decoder when performing the alignment with $\mathrm{score}()$, since there is no decoder. So, in self-attention, the alignment vector $\alpha$ indicates the \textit{similarity of each input word with respect to the optimal word (on average)}, while in seq2seq attention, $\alpha$ indicates the \textit{relevance of each source word in generating the next element of the target sentence}.

\subsubsection{Hierarchical attention}
\begin{figure}[H]
\centering
\includegraphics[width=0.95\textwidth]{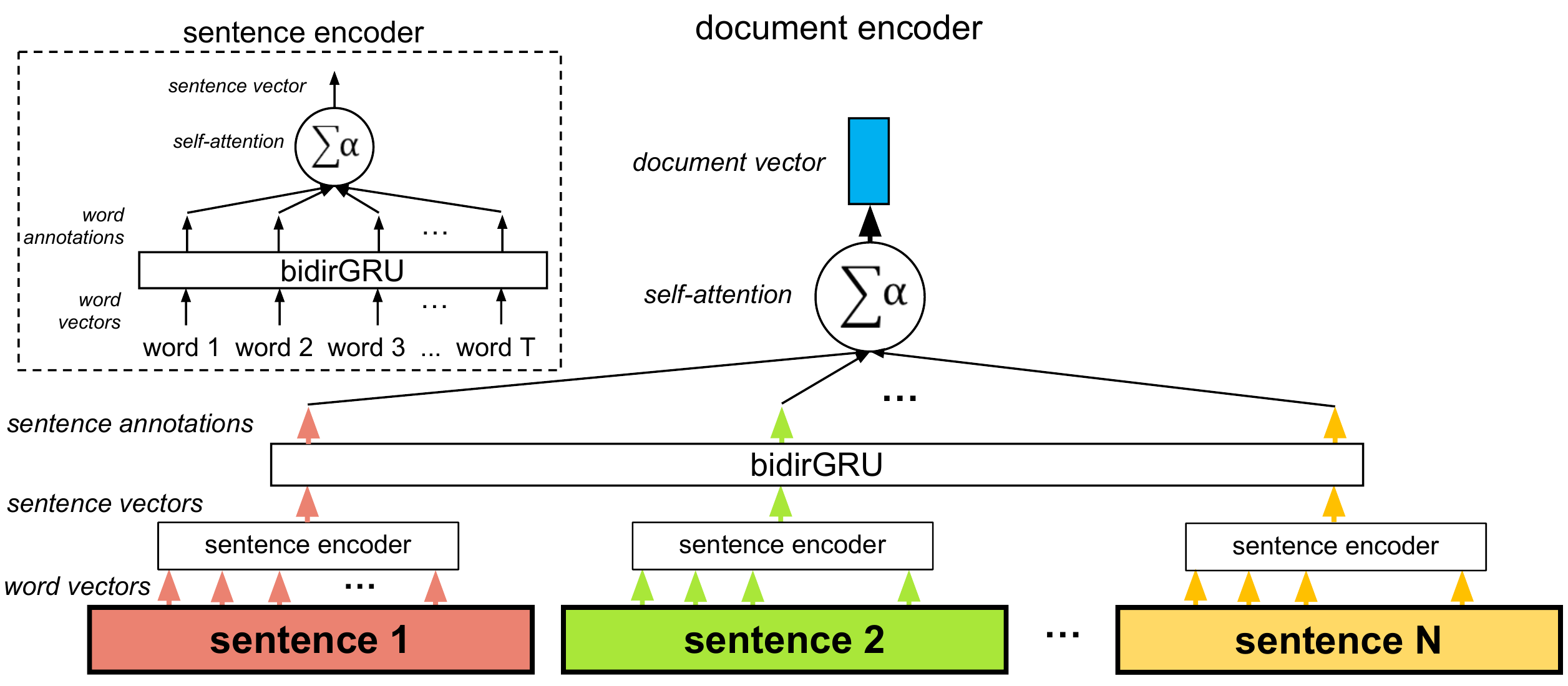}
\captionsetup{size=small}
\caption{Hierarchical attention architecture \cite{han}.}\label{fig:han}
\end{figure}

A simple, good example of how self-attention can be useful in practice is provided by the architecture illustrated in Fig. \ref{fig:han}. In this architecture, the self-attention mechanism comes into play twice: at the word level, and at the sentence level. This approach makes sense for two reasons: first, it matches the natural hierarchical structure of documents (words $\rightarrow$ sentences $\rightarrow$ document). Second, in computing the encoding of the document, it allows the model to first determine which words are important in each sentence, and then, which sentences are important overall. Through being able to re-weigh the word attentional coefficients by the sentence attentional coefficients, the model captures the fact that a given instance of word may be very important when found in a given sentence, but another instance of the same word may not be that important when found in another sentence.

\begin{thebibliography}{9}

\bibitem{att_bahdanau}
Bahdanau, Dzmitry, Kyunghyun Cho, and Yoshua Bengio. "Neural machine translation by jointly learning to align and translate." arXiv preprint arXiv:1409.0473 (2014).

\bibitem{gru}
Cho, K., Van Merriënboer, B., Gulcehre, C., Bahdanau, D., Bougares, F., Schwenk, H., Bengio, Y. (2014). Learning phrase representations using RNN encoder-decoder for statistical machine translation. arXiv preprint arXiv:1406.1078.

\bibitem{chopra}
Chopra, Sumit, Raia Hadsell, and Yann LeCun. "Learning a similarity metric discriminatively, with application to face verification." Computer Vision and Pattern Recognition, 2005. CVPR 2005. IEEE Computer Society Conference on. Vol. 1. IEEE, 2005.

\bibitem{gru_emp}
Chung, J., Gulcehre, C., Cho, K., \& Bengio, Y. (2014). Empirical evaluation of gated recurrent neural networks on sequence modeling. arXiv preprint arXiv:1412.3555.

\bibitem{infersent}
Conneau, A., Kiela, D., Schwenk, H., Barrault, L., \& Bordes, A. (2017). Supervised learning of universal sentence representations from natural language inference data. arXiv preprint arXiv:1705.02364.

\bibitem{elman1990finding}
Elman, J. L. (1990). Finding structure in time. Cognitive Science, 14:2, 179-211.

\bibitem{beam}
Freitag, Markus, and Yaser Al-Onaizan. "Beam search strategies for neural machine translation." arXiv preprint arXiv:1702.01806 (2017).

\bibitem{goldberg2016primer}
Goldberg, Y. (2015). A primer on neural network models for natural language processing. Journal of Artificial Intelligence Research,  57, 345-420.

\bibitem{graves_2013}
Graves, A. (2013). Generating sequences with recurrent neural networks. arXiv preprint arXiv:1308.0850.

\bibitem{lstm_odyssey}
Greff, Klaus, et al. "LSTM: A search space odyssey." IEEE transactions on neural networks and learning systems 28.10 (2017): 2222-2232.

\bibitem{1997lstm}
Hochreiter, S., Schmidhuber, J. (1997). Long short-term memory. Neural computation, 9(8), 1735

\bibitem{hubel1962receptive}
Hubel, David H., and Torsten N. Wiesel (1962). Receptive fields, binocular interaction and functional architecture in the cat's visual cortex. The Journal of physiology 160.1:106-154.

\bibitem{JohnsonZhang}
Johnson, R., Zhang, T. (2015). Effective Use of Word Order for Text Categorization with Convolutional Neural Networks. To Appear: NAACL-2015, (2011).

\bibitem{kim2014}
Kim, Y. (2014). Convolutional Neural Networks for Sentence Classification. Proceedings of the 2014 Conference on Empirical Methods in Natural Language Processing (EMNLP 2014), 1746–1751.

\bibitem{krizhevsky}
Krizhevsky, Alex, Ilya Sutskever, and Geoffrey E. Hinton. "Imagenet classification with deep convolutional neural networks." Advances in neural information processing systems. 2012.

\bibitem{lecun1998}
LeCun, Y., Bottou, L., Bengio, Y., and Haffner, P. (1998). Gradient-based learning applied to document recognition. Proceedings of the IEEE, 86(11), 2278-2324.

\bibitem{lin2015visualizing}
Li, J., Chen, X., Hovy, E., and Jurafsky, D. (2015). Visualizing and understanding neural models in nlp. arXiv preprint arXiv:1506.01066.

\bibitem{self-attentive}
Lin, Zhouhan, et al. "A structured self-attentive sentence embedding." arXiv preprint arXiv:1703.03130 (2017).

\bibitem{lipton}
Lipton, Zachary C., John Berkowitz, and Charles Elkan. "A critical review of recurrent neural networks for sequence learning." arXiv preprint arXiv:1506.00019 (2015).

\bibitem{att_luong}
Luong, Minh-Thang, Hieu Pham, and Christopher D. Manning. "Effective approaches to attention-based neural machine translation." arXiv preprint arXiv:1508.04025 (2015).

\bibitem{att_summ}
Rush, Alexander M., Sumit Chopra, and Jason Weston. "A neural attention model for abstractive sentence summarization." arXiv preprint arXiv:1509.00685 (2015).

\bibitem{simonyan2013deep}
Simonyan, K., Vedaldi, A., and Zisserman, A. (2013). Simonyan, Karen, Andrea Vedaldi, and Andrew Zisserman. "Deep inside convolutional networks: Visualising image classification models and saliency maps." arXiv preprint arXiv:1312.6034 (2013). arXiv preprint arXiv:1312.6034.

\bibitem{springenberg2014striving}
Springenberg, Jost Tobias, et al. "Striving for simplicity: The all convolutional net." arXiv preprint arXiv:1412.6806 (2014).

\bibitem{seq-to-seq}
Sutskever, Ilya, Oriol Vinyals, and Quoc V. Le. "Sequence to sequence learning with neural networks." Advances in neural information processing systems. 2014.

\bibitem{show_attend_tell}
Xu, K., Ba, J., Kiros, R., Cho, K., Courville, A., Salakhudinov, R., ... \& Bengio, Y. (2015, June). Show, attend and tell: Neural image caption generation with visual attention. In International Conference on Machine Learning (pp. 2048-2057).


\bibitem{Zhang2015practicioner}
Zhang, Ye, and Byron Wallace. "A sensitivity analysis of (and practitioners' guide to) convolutional neural networks for sentence classification." arXiv preprint arXiv:1510.03820 (2015).

\bibitem{han}
Yang, Zichao, et al. "Hierarchical attention networks for document classification." Proceedings of the 2016 Conference of the North American Chapter of the Association for Computational Linguistics: Human Language Technologies. 2016.

\end{thebibliography}

\end{document}